\def\year{2020}\relax
\documentclass[letterpaper]{article} 
\usepackage{aaai20}  
\usepackage{times}  
\usepackage{helvet} 
\usepackage{courier}  
\usepackage{amsmath,,amsfonts,amssymb}
\usepackage{array,multirow,graphicx}
\usepackage{float}
\usepackage{CJKutf8}
\usepackage{bm}
\usepackage{natbib}
\usepackage[hyphens]{url}  
\usepackage{graphicx} 
\usepackage{amsthm}
\usepackage{multirow}
\usepackage{graphicx}
\usepackage{booktabs}
\usepackage{graphicx}  
\newcommand{\approach}[1]{approach}

\usepackage{etoolbox}
\makeatletter
\patchcmd{\maketitle}{\@copyrightspace}{}{}{}
\makeatother

\title{Speech2Slot: An End-to-End Knowledge-based Slot Filling from Speech}

\author{
\Large \textbf{Pengwei Wang, Xin Ye, Xiaohuan Zhou, Jinghui Xie, Hao Wang} \\
Alibaba Group \\
\{hoverwang.wpw,sunny.yx,shiyi.zxh,jinghui.xjh,qiao.wh\}@alibaba-inc.com
}

\begin{document}
 
\maketitle

\begin{abstract}
In contrast to conventional pipeline Spoken Language Understanding (SLU) which consists of automatic speech recognition (ASR) and natural language understanding (NLU), end-to-end SLU infers the semantic meaning directly from speech and overcomes the error propagation caused by ASR.
End-to-end slot filling (SF) from speech is an essential component of end-to-end SLU, and is usually regarded as a sequence-to-sequence generation problem, heavily relied on the performance of language model of ASR.
However, it is hard to generate a correct slot when the slot is out-of-vovabulary (OOV) in training data, especially when a slot is an anti-linguistic entity without grammatical rule.
Inspired by object detection in computer vision that is to detect the object from an image, we consider SF as the task of slot detection from speech.
In this paper, we formulate the SF task as a matching task and propose an end-to-end knowledge-based SF model, named Speech-to-Slot (Speech2Slot), to leverage knowledge to detect the boundary of a slot from the speech.
We also release a large-scale dataset of Chinese speech for slot filling, containing more than 830,000 samples.
The experiments show that our approach is markedly superior to the conventional pipeline SLU approach, and outperforms the state-of-the-art end-to-end SF approach with \textbf{12.51\%} accuracy improvement.

\end{abstract}
\noindent\textbf{Index Terms}: Speech2Slot, knowledge, object detection, out-of-vocabulary, anti-linguistic

\section{Introduction}
Spoken Language Understanding (SLU) has attracted much attention in recent years with the rapidly growing demand for voice assistants such as Siri, Cortana, Alexa, and Google Home etc.  \cite{Tur:2012,Xu:2014,Yao:2014,Siddhant:2019,Zhao:2019}. 
The conventional pipeline approaches typically consist of two parts: automatic speech recognition (ASR), which converts speech into the underlying text, and natural language understanding (NLU), which learns semantics from the converted text input \cite{Coucke:2018,Gorin:1997,Mesnil:2015}.
The NLU task mainly contains domain classification, intent classification, and slot filling.
The major problem of such approaches is that NLU suffers from the upstrem ASR errors, which set an accuracy upper bound of the entire system.
ASR usually confuses similar sounds and leads to errors, especially for the sounds of slots.
For example, ``\emph{I want to listen to Maria}'' is an user query in a music play scene.
The query pattern (``\emph{I want to listen to ...}'') can be exactly recognized by current ASR system.
However, the music name (``\emph{Maria}'') is always wrongly recognized, such as ``\emph{Marine}'', ``\emph{Mariya}'' or ``\emph{Merian}'' etc.
The ASR error mainly focuses on the misrecognition of slots.
Thus, in this paper we focus on slot filling (SF) from speech.

Some end-to-end SLU approaches, which directly infer semantic meaning from speech, have been proposed to eliminate the error propagation problem \cite{Chen:2018,Haghani:2018,Serdyuk:2019,Lugosch1:2019,Price:2020,Qian:2021}.
Most of these studies focus on end-to-end domain/intent classification by leveraging a unified end-to-end neural network to learn the sentence-level representation of the input speech.
To apply the end-to-end modeling approach in SF task, the cross-modal encoder-decoder modeling approaches \cite{Haghani:2018,Tomashenko:2019,Rao:2020,Lai:2020,Kuo:2020,Pelloin:2021,Ghannay:2018,Yadav:2020} are proposed, where the speech is used as the input to generate the slots.
In \cite{Ghannay:2018,Yadav:2020}, the slots are directly decoded in the encoder-decoder network by adding some special symbols to denote the start and end of the slot.
To further eliminate the error propagation from ASR, jointly training the ASR and SF components in a manner of sequence-to-sequence mapping \cite{Haghani:2018,Tomashenko:2019,Rao:2020,Lai:2020,Kuo:2020,Pelloin:2021} is proposed.

In previous studies, SF is formulated as a conditional sequence to sequence generation task, given the contextual embeddings of input speech.
The slot decoding in a generation manner relies heavily on the performance of language model (LM).
Further, it is found that the number of slots in most vertical domains, such as locations, music and video etc., are large-scale. 
It is impossible to cover the entire slots in the training data.
In real production environments, many new slots would appear in a speech, which are named as Out-of-vocabulary (OOV) slots.
The LM will make mistakes when an OOV slot is in a speech, e.g. music name ``\emph{Maria}''.
Moreover, anti-linguistic character, containing no grammatical rule, is also extremely common in slots.
The LM prefers to generate a wrong slot with higher probability, rather than the correct anti-linguistic slot.
For example, the movie ``\emph{knight and day}'' is always wrongly recognized as ``\emph{night and day}" by LM.

Matching an OOV or anti-linguistic slot with an input speech seems more feasible than directly generating the slot from the speech.
Therefore, we formulate the SF task as a matching task to address the OOV and anti-linguistic challenges.
Obtaining accurate boundaries of sub-word from speech in a matching task is nontrivial.
Inspired by object detection in computer vision that is to detect the object from an image, we consider SF as the task of slot detection from speech.
In this paper, we propose an end-to-end knowledge-based SF model, named Speech-to-Slot (Speech2Slot), to leverage knowledge to detect the boundary of a slot from the speech.
The knowledge refers to the entity database.
Specifically, the input of the Speech2Slot model is phoneme posterior and entity database, and the output is the phone sequence of the slot.
The phoneme posterior is generated by an acoustic model (AM).
The entity database can be regraded as a close set.
Thus, our goal is to select the correct slot from the entity database according to the input speech.
Specially, the entity database is firstly used to build a trie-tree.
Then, the trie-tree detects the start timestamp and end timestamp of an entity in the phoneme posterior of a speech.
Finally, a bridge layer is proposed to decode the slot phone sequence from entity database according to the detected phoneme fragment.


To the best of our knowledge, almost all of the existing SLU datasets are in English and small in scale.
Thus, we release a large-scale Chinese speech-to-slot dataset in the domain of voice navigation, which contains  820,000 training samples and 12,000 testing samples.
We compare the proposed Speech2Slot model to a conventional pipeline SLU approach and a state-of-the-art end-to-end SLU approach.
The experiments show that our approach outperforms the conventional pipeline SLU approach and the end-to-end SF approach with over 46.44\% and 12.51\% accuracy improvement separately.
Especially in the OOV slot and anti-linguistic slot, the proposed Speech2Slot model achieves significantly improvement over the conventional pipeline SLU approach and the end-to-end SF approach.


\section{Related Work}
\label{sec:related_work}
There is plenty of work on conventional pipeline Spoken Language Understanding (SLU) \cite{Coucke:2018,Gorin:1997,Mesnil:2015,Devlin:2019,Nadeau:2007,Peters:2018,Baevski:2019}.
In the slot filling part of SLU, most studies \cite{Jiang:2019,Strakova:2019,Parada:2011,Cohn:2019,Yan:2019} focus on extracting slots from the textual documents.
The ASR error easily propagates to the SF model.
In past few year, end-to-end SLU models are proposed to address the error propagate problem \cite{Qian:2017,Chen:2018,Haghani:2018,Serdyuk:2019,Lugosch1:2019,Price:2020,Huang2020LeveragingUT,Qian:2021}.
The end-to-end SLU models leverage a unified neural network (e.g. RNN \cite{Serdyuk:2019} or transformer \cite{Wang:2020}) to replace the ASR model and NLU model, where raw waveforms or the acoustic features are directly used as the inputs to infer the NLU result.
However, most of these studies aim to get the sentence-level representation of the input speech, which can only be used in the domain classification and intent classification.
To apply the end-to-end modeling approach in slot filling task, the cross-modal encoder-decoder modeling approaches have been proposed
\cite{Haghani:2018,Tomashenko:2019,Rao:2020,Lai:2020,Kuo:2020,Pelloin:2021,Ghannay:2018,Yadav:2020}.
In \cite{Ghannay:2018,Yadav:2020,Chuang:2020}, the slots are directly decoded in the encoder-decoder network by adding some special symbols to denote the start and end of the slot.
However, these models need the alignment between the speech segment and the transcript word token, which is an expensive and time-consuming process.
To further eliminate the ASR error propagation, jointly training the ASR and SF components in a manner of sequence-to-sequence mapping \cite{Haghani:2018,Tomashenko:2019,Rao:2020,Lai:2020,Kuo:2020,Pelloin:2021} is proposed.
To sum up, the previous end-to-end approaches still regard the SF task as a generating task, where the slot decoding relies heavily on the performance of language model.
The main challenges, OOV and the anti-linguistic character, are not addressed very well in generation based approaches.
Thus, we propose to formulate the SF task as a matching task to address the two challenges.

\section{Speech-to-Slot Model}
\label{sec:model}
\begin{figure*}[!ht]
	\centering
	\includegraphics[scale=0.57]{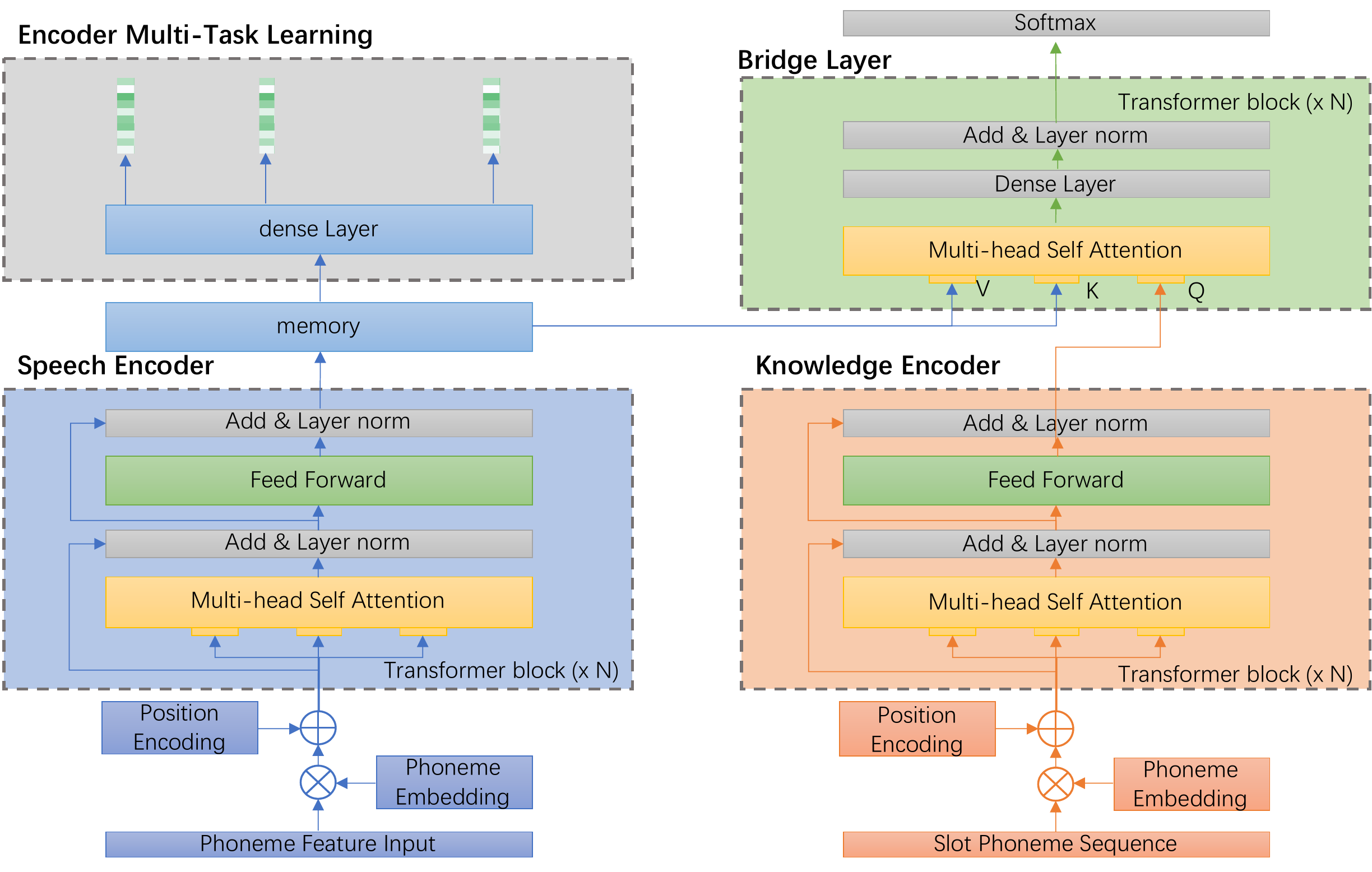}
	\caption{
		\label{fig:phone2slot} Speech-to-Slot Model}
\end{figure*}

The entire end-to-end slot filling from speech consists of two major components: AM and Speech2Slot.
The AM is trained with a CTC loss over the ground-truth phoneme output.
The phoneme logits are further fed to a softmax layer to get the posterior distribution over phonemes at each frame.
The Speech2Slot model takes the phoneme posterior distribution as its input.
Such implementation makes Speech2Slot independent from the AM implementation conditioned on the phoneme posterior. 
Therefore, Speech2Slot is less sensitive to the choice of the AM implementation as long as it provides proper phoneme posterior with comparable accuracy.
In this paper, we use a open source code\footnote{https://github.com/audier/DeepSpeechRecognition} to train the AM. 
As shown in Figure~\ref{fig:phone2slot}, our proposed Speech2Slot model consists of speech encoder, knowledge encoder, and bridge layer.

\subsection{Speech Encoder}
To obtain the speech representation, the input phoneme posterior feature is encoded by the speech encoder.
We employ transformer encoder network as the speech encoder, since it has been proven effective in almost all NLU tasks \cite{Devlin:2019}. 
The input of the transformer encoder network is the phoneme posterior.
The output of the last transformer block is fed into a bridge layer and an encoder multi-task learning layer.
A relative position embedding and phoneme embedding is used to capture the phoneme posterior semantic and position information.
To minimize the overfitting effect, we build a multi-task learning on the encoder. 
We follow \cite{Wang:2020} to randomly mask 10\%-15\% of the frames from the input phoneme posteriors.
The masked frames are filled with a default padding frame where the probability of the placeholder phoneme ``[PAD]'' equals 1.
At the speech encoder hidden vectors of the masked frames, we add a dense layer to predict the masked frame.
The objective function is the cross entropy between the original phoneme posterior frame and the predicted ones.
\subsection{Knowledge Encoder}
The knowledge encoder is used to obtain the representation of the entire slots in the knowledge base.
We also employ a transformer encoder network as the knowledge encoder.
The input of the knowledge encoder is the slot phoneme sequence.
The output of the last transformer block is fed into a bridge layer.
The knowledge encoder is mainly in charge of remembering the entire slots.
Thus, the knowledge encoder can be trained on the entire slots in advance.
The parameters of the trained knowledge encoder can be fixed or fine-tuned in the training process of Speech2Slot.
\subsection{Bridge Layer}
The birdge layer is used to detect the slot boundary (i.e. start timestamp and end timestamp of a slot) from the input phoneme posterior according to the slot representation from knowledge encoder.
By removing the ResNet mechanism from knowledge encoder to the bridge layer, the slot extraction information just comes from the speech representation generating from speech encoder, which is essential difference with existing end-to-end SF approaches.
This mechanism can completely solve the over-reliance on language model, and address the OOV and anti-linguistic problems.
The birdge layer uses a transformer structure by removing the ResNet with knowledge encoder.
The memory of the speech encoder is served as the value input (V) and key input (K).
The memory of the knowledge encoder is served as the query input (Q).
At the hidden vectors of the last block, we add a softmax to predict the slot phoneme sequence.
\subsection{Testing phase}
In the testing phase, the entire slots are firstly used to build a trie-tree.
Then, the trie-tree detects the start timestamp and end timestamp of a slot in the speech.
Finally, the most matched slot phone sequence with the detected speech fragment is the output of Speech2Slot model.
If there is no slot in an input speech, the confidence coefficient of the Speech2Slot model can be used to filter these cases.
\section{Dataset}
\label{sec:dataset}
\begin{CJK}{UTF8}{gbsn}
This section describes the preparation of a Chinese dataset of voice navigation, named Voice Navigation in Chinese.
First, we collect more than 830,000 place names in China, such as ``\emph{故宫}''(\emph{The Palace Museum}), ``\emph{八达岭长城}''(\emph{Great Wall on Badaling}),  ``\emph{积水潭医院}''(\emph{Jishuitan Hospital}) etc.
To generate the navigation queries, we also collect more than 25 query patterns, as shown in Table~\ref{tab:query_pattern}. 
We fill out the query pattern with places to generate the query.
\begin{table}[h]
\caption{Query patterns for Voice Navigation scene}
\label{tab:query_pattern}       
\centering
\begin{tabular}{l|l}
\hline\noalign{\smallskip}
Chinese & Corresponding English \\
\noalign{\smallskip}\hline\noalign{\smallskip}
导航去[\emph{地点}] & Navigation to [\emph{location}] \\
我要去[\emph{地点}] & I want to go to [\emph{location}] \\
去[\emph{地点}]的路线 & The route to [\emph{location}] \\
去[\emph{地点}] & Go to [\emph{location}] \\
搜索[\emph{地点}] & Search [\emph{location}] \\
\noalign{\smallskip}\hline
\end{tabular}
\end{table}
\end{CJK}

Next, we use a TTS tool\footnote{https://www.xfyun.cn/services/offline\_tts} to generate the speech file by reading the generated query. 
The TTS tool provides several variant articulation types, such as men, women, background music, echo and underwater etc.
The variant articulation types can increase the data abundance greatly.
In addition, to build a real-person testing data, we invite 5 persons to read 2000 cases.
For the testing data, we call the data generated by TTS as TTS data, and the data generated by real person as human-read data.
Finally, we generate 820,000 speech-slot pairs for training and 12,000 speech-slot pairs for testing. 
Further, we would verify the effect on the OOV and Non-OOV dataset.
Half of the slots in testing dataset do not appear in training datase.
The details are listed in Table~\ref{tab:dataset}.
We have released the entire training and testing data\footnote{https://1drv.ms/u/s!AmYoFTLtyCO\_k49SFLCJOneZBY\\CWCA?e=2GgEcS}.

\begin{table}
\renewcommand{\multirowsetup}{\centering}
\caption{The statistic of Voice Navigation dataset}
\label{tab:dataset}       
\centering
\begin{tabular}{c|c|c|c|c}
\hline\noalign{\smallskip}
\multirow{3}{*}{Train}
&\multicolumn{4}{c}{Test}\\
\cmidrule(r){2-5}
&\multicolumn{2}{c|}{TTS}&\multicolumn{2}{c}{Human-read}\\
\cmidrule(r){2-5}
&OOV&Non-OOV&OOV&Non-OOV\\
\noalign{\smallskip}\hline\noalign{\smallskip}
820,000 & 5000 & 5000 & 1000 & 1000 \\
\noalign{\smallskip}\hline
\end{tabular}
\end{table}


\section{Experiment}
\label{sec:experiment}
\subsection{Experiment Settings}
 \noindent\textbf{Hyper-parameters}. The number of transformer layers in speech encoder and knowledge encoder is set to 4. 
The number of transformer layers in bridge layer is set to 2.
In all of the transformer layers, the hidden size is set to 576, the intermediate size is set to 1600 and the number of heads of self-attention is set to 8. 
The learning rate is set to $3*10^{-5}$ and the dropout rate is set to 0.1. 
We separately set the max input of the speech encoder and knowledge encoder to be 40 and 10.

\noindent\textbf{Environment}. All models are implemented with Tensorflow 1.12. We conduct the Speech2Slot experiment on 8 Nvidia-Tesla P-100 GPUs for approximately 30 minutes for 20 epochs.
\subsection{Experiment Methods}
\textbf{AM+LM} \cite{Yan:2019}. We implement a conventional pipeline SLU approach by cascading the ASR model (AM+LM) with a SF model \cite{Yan:2019}. 
The SF is implemented by a transformer stacked by a CRF.
We denote this method as AM+LM.
 
\noindent\textbf{AM+S2Q} \cite{Chan:2016}. We also implement an encoder-decoder ASR \cite{Chan:2016} to decode the query.
The transformer is used in encoder and decoder.
A text-based SF model, same with AM+LM method, is used to extract the slot. We denote this method as AM+S2Q.

\noindent\textbf{AM+SF} \cite{Pelloin:2021}. To verify the performance of the end-to-end SF model, we implement an end-to-end SF model in \cite{Pelloin:2021}, which is denoted as AM+SF.

\noindent\textbf{AM+S2S}: Our proposed Speech2Slot model is denoted as AM+S2S in following tables.
\subsection{Experiment Results}
We compare different approaches on the Voice Navigation dataset.
The result of experimenting on the TTS testing data is shown in Table~\ref{tab:res_whole_data}.
To valid the AM effect on Speech2Slot model, we also compare the different AM model results.
We train the AM with our release domain-specific training data (Domain-AM) and open-source ASR training data\footnote{http://www.openslr.org/18/} (General-AM) respectively.
Most of the slots in testing data are anti-linguistic.
As we can see from Table~\ref{tab:res_whole_data}, our proposed Speech2Slot model significantly outperforms the AM+LM by a large margin in both of the AM models.
In addition, Speech2Slot model also outperforms the end-to-end SF approach and the S2Q approach.
Especially, the Speech2Slot can also maintain a great performance in OOV dataset. 
From the experiments, we observe that the proposed Speech2Slot can efficiently eliminate the error propagation and address the OOV and the anti-linguistic challenges regardless of the different AM phoneme posteriors.
This also verifies that the matching manner in end-to-end SF is more effective than the generating manner.
\begin{table}[h]
\caption{Comparison of SF accuracy between different approaches on the TTS dataset}
\label{tab:res_whole_data}       
\centering
\begin{tabular}{l|c|c|c}
\hline\noalign{\smallskip}
Model & Entire & Non-OOV & OOV \\
\noalign{\smallskip}\hline\noalign{\smallskip}
Domain-AM+LM & 37.42\% & 51.66\% & 23.18\%\\
Domain-AM+S2Q & 71.23\% & 80.52\% & 61.94\%\\
Domain-AM+SF & 71.35\% & 80.58\% & 61.90\%\\
Domain-AM+S2S & \textbf{83.86\%} & \textbf{ 85.82\%} &  \textbf{81.90\%}\\
General-AM+LM & 1.18\% & 1.34\% & 1.02\%\\
General-AM+S2Q & 43.95\% & 53.64\% & 34.26\%\\
General-AM+SF & 44.01\% & 53.67\% & 34.53\%\\
General-AM+S2S & \textbf{72.99\%} & \textbf{76.08\%} & \textbf{69.90\%}\\
\noalign{\smallskip}\hline
\end{tabular}
\end{table}


We further experiment on the human-read test data, and verify the effect of the Speech2Slot in real person speech.
Due to the fact that the AM trained by TTS data is not suitable for obtaining the phoneme posterior of real human speech, we only use the general AM in this experiment.
As shown in Table~\ref{tab:res_human_data}, the accuracy of the all models are extremely low.
This is because the quality of the phoneme posterior generated from the general-AM model is low for real person speech.
Even so the Speech2Slot model significantly outperforms the baselines.
Especially in OOV dataset, the Speech2Slot can still achieve 16.5\% accuracy.
From the experiments, we can see that the Speech2Slot can get a better performance in real production environments compared with other baselines.
We believe the performance of Speech2Slot can be potentially improved if a high quality phoneme posterior is provided.
\begin{table}[h]
\caption{Comparison of SF accuracy between different approaches on the human-read dataset}
\label{tab:res_human_data}       
\centering
\begin{tabular}{l|c|c|c}
\hline\noalign{\smallskip}
Model & Entire & Non-OOV & OOV \\
\noalign{\smallskip}\hline\noalign{\smallskip}
General-AM+LM & 0.50\% & 1.00\% & 0.01\%\\
General-AM+S2Q & 6.16\% & 11.30\% & 0.92\%\\
General-AM+SF & 6.18\% & 11.40\% & 0.95\%\\
General-AM+S2S & \textbf{23.60\%} & \textbf{30.7\%} & \textbf{16.50\%}\\
\noalign{\smallskip}\hline
\end{tabular}
\end{table}


\section{Conclusion}
In this paper, an end-to-end knowledge-based slot filling model, named Speech2Slot, is proposed.
We are the first to formulate the slot filling as a matching task instead of a generation task.
The slot is extracted by matching the detected slot fragment in speech with the entity database.
In contrast to the generation manner, the matching manner can efficiently address the OOV and anti-linguistic problems.
In addition, we release a large-scale Chinese speech-to-slot dataset in the domain of voice navigation.
The experiment results show that our proposed Speech2Slot can significantly outperform the pipeline SLU approach and the state-of-the-art end-to-end SF approach. 
Especially, Speech2Slot almost has the same performance whatever it meets OOV or not.
In future work, to eliminate the error caused by AM, more raw speech features will be used to extract slots.
And multi-slots filling is also an important and practical research direction.

\bibliography{main}
\bibliographystyle{aaai}

\end{document}